\documentclass{article}

\usepackage[preprint]{neurips_2025}

\usepackage[utf8]{inputenc} 
\usepackage[T1]{fontenc}    
\usepackage{hyperref}       
\usepackage{url}            
\usepackage{booktabs}       
\usepackage{amsfonts}       
\usepackage{nicefrac}       
\usepackage{microtype}      
\usepackage[dvipsnames]{xcolor}
\usepackage[many]{tcolorbox}    
\usepackage{enumitem}

\usepackage{amsmath}
\usepackage{amssymb}
\usepackage{mathtools}
\usepackage{amsthm}
\usepackage{framed}

\usepackage[capitalize,noabbrev]{cleveref}
\crefformat{footnote}{#2\footnotemark[#1]#3}

\usepackage{amsmath,amsfonts,bm}
\usepackage{siunitx}

\usepackage{enumitem}

\def\eqref#1{equation~\ref{#1}}

\def\1{\bm{1}}

\DeclareMathAlphabet{\mathsfit}{\encodingdefault}{\sfdefault}{m}{sl}
\SetMathAlphabet{\mathsfit}{bold}{\encodingdefault}{\sfdefault}{bx}{n}

\newcommand{\2}{^{2}}

\DeclareMathSymbol{\shortminus}{\mathbin}{AMSa}{"39}

\makeatletter
\def\section{\@startsection {section}{1}{\z@}{-1ex plus -0.2ex minus 0.2ex}{0.1ex plus 0.1ex minus 0.1ex}{\large\bf\raggedright}}
\def\paragraph{\@startsection{paragraph}{4}{\z@}{.1ex plus 0ex minus .1ex}{-1em}{\normalsize\bf}}
\makeatother
\theoremstyle{plain}
\newtheorem{theorem}{Theorem}[section]
\newtheorem{proposition}[theorem]{Proposition}

\theoremstyle{definition}

\theoremstyle{remark}

\usepackage{amsthm}
\usepackage[dvipsnames]{xcolor}
\theoremstyle{definition}
\theoremstyle{plain}

\def\bw{\mathbf{w}}
\def\bgamma{\boldsymbol{\gamma}}

\def\distrib#1{^{[#1]}}
\def\pre{\distrib{\mathrm{mix}}}
\def\final{\distrib{i}}
\def\bal{\distrib{bal}}
\def\j{\distrib{j}}

\def\adja{\distrib{\mathrm{adja}}}

\def\cor{\mathrm{Cor}}

\def\1{\distrib{1}}
\def\2{\distrib{2}}
\def\3{\distrib{3}}
\def\4{\distrib{4}}
\usepackage{multirow}
\usepackage{graphicx}
\usepackage{booktabs, tabularx} 
\usepackage[export]{adjustbox}   
\usepackage{eurosym}             

\makeatletter
\def\section{\@startsection {section}{1}{\z@}{-1ex plus -0.2ex minus 0.2ex}{0.1ex plus 0.1ex minus 0.1ex}{\large\bf\raggedright}}
\def\paragraph{\@startsection{paragraph}{4}{\z@}{.1ex plus 0ex minus .1ex}{-1em}{\normalsize\bf}}
\makeatother
\newtcolorbox{boxA}{
    fontupper = \it,
    boxrule = .5pt,
    colframe = black 
}

\title{These Are Not All the Features You Are Looking For: \\ A Fundamental Bottleneck in Supervised Pretraining}

\author{%
  Xingyu (Alice) Yang \\
  Fundamental AI Research (FAIR) at Meta \\
  New York, United States \\
  \texttt{axyang@meta.com}
  \And 
  Jianyu Zhang \\
  New York University \\
  New York, United States \\
  \texttt{jianyu@nyu.edu}
  \And 
  L\'eon Bottou \\
  Fundamental AI Research (FAIR) at Meta \\
  New York, United States \\
  \texttt{leon@bottou.org}
  }

\begin{document}
\maketitle
\begin{abstract}
Transfer learning is widely used to adapt large pretrained models to new tasks with only a small amount of new data. However, a challenge persists -- the features from the original task often do not fully cover what is needed for unseen data, especially when the relatedness of tasks is not clear.
Since deep learning models tend to learn very sparse representations, they retain only the minimal features required for the initial training while discarding potentially ones for downstream transfer. 
A theoretical framework developed in this work demonstrates that such pretraining captures inconsistent aspects of the data distribution, therefore, inducing transfer bias.
To address this limitation, we propose an inexpensive ensembling strategy that aggregates multiple models to generate richer feature representations. On ResNet, this approach yields a $9\%$ improvement in transfer accuracy without incurring extra pretraining cost. 
We also present empirical evidence from a range of deep learning studies, confirming that the phenomenon is pervasive across modern deep learning architectures. 
These results suggests that relying solely on large pretrained networks is not always the most effective way to improve model generalization. Instead, fostering richer, more diverse representations -- e.g. - through model ensembles -- can substantially enhance transfer learning performance.
\end{abstract}

\section{Introduction}

Transfer learning has become a cornerstone of modern machine learning and artificial intelligence \citep{tr-bottou-2011,collobert-2011,oquab-2014,bommasani-2021}. In its simplest form--supervised transfer learning--a large neural network is first trained on inexpensive and diverse dataset that is related in some way to the target task. The network is then fine-tuned on a much smaller, task-specific dataset. Since the fine-tuning dataset is traditionally too small too small to train a network of that size from scratch, the hope is that the representations learned during the pretraining phase will be useful during the adaptation phase  \citep{pacs}. 

A key challenge is ensure that transferred features are sufficient for unseen data. When pre-training data is unrelated to the target task, the adapted model can underperform compared with a model trained directly on ample task-specific data.

The folklore around foundation model suggests that training an enormous model on "everything" will give it essentially all useful features \cite{bommasani-2021}. We ask a more modest question: \textit{if a model is pretrained on a mixture of data that includes the target task, can the fine-tuned model match the performance of a model trained directly on a large, task-specific dataset?
}

Answering this question informs whether researchers should invest in ever larger, all-purpose foundation models or in collecting task-specific datasets for training smaller specialized models.

Our findings are negative : even under favorable conditions--a pretraining mixture that contains the target task--a fine-tuned model will still fall short of a model trained directly on ample task data. In other words, pretraining on mixed data does not guarantee that fine-tuning will achieve parity with task-specific training.

\begin{framed} 
If a model pretrained on mixed data that includes the target task cannot consistently match a model trained directly on that task, then we should not expect good performance when we have little knowledge of how the pretraining data relates to the target task.
\end{framed}

We uncover a fundamental bottleneck caused by the sparsity bias of deep learning networks. During pretraining, a model learns only the minimal set of features required for that task, discarding other features that may be essential for downstream problems. As a result, crucial information is lost and cannot be recovered by simply fine-tuning on a small target dataset -- even when the target task is represented in the pretraining data.  Once a feature is encoded, a network often fails to learn competing features, which hampers adaptation to new or previously seen tasks.

Although previously overlooked, this phenomenon is pervasive. We illustrate it with a simple theoretical counterexample and discuss empirical studies that reveal consistent feature loss. Factors such as class-imbalance and data-shuffling order further influence which features are retained. 

To address this limitation, we propose an inexpensive  strategy that improves generalization to unseen distributions without additional cost. For example, on ResNet50, this approach yields a $9\%$ improvement in transfer accuracy without incurring extra pretraining cost. 

The paper is organized as follows:
\begin{itemize}
\item \textbf{Section 2} formalizes the effectiveness of transfer.
\item \textbf{Section 3} presents the counterexample.
\item \textbf{Section 4} discusses the bottleneck, supports it with empirical evidence, and introduces rich representations as a solution.

\item \textbf{Section 5} describes a method for constructing such representations and demonstrates its benefits on unseen data..
\end{itemize}

\section{Problem Statement}
\label{problem statement}
Let $\{ P\j \}_j$ denote a set of probability distributions over (X,Y), each corresponding to data from a distinct population. 
Consider a \textbf{non-trivial} mixture 
$$
P\pre(X,Y)= \sum_{j=1}^n \lambda_j  P\j(X,Y), \quad 
 \lambda_j>0,~ 
 \sum_{j=1}^{n} \lambda_j = 1.
$$
Assume a neural network is pretrained on samples drawn from $P\pre$.
Let $P\final$ be an arbitrary target distribution from the same family.
If we have enough data, we also could train a network directly on $P\final$.\\  \\
\textbf{Question:} 
\emph{Do the feature representations learned from pretraining on $P\pre$ achieve performance on $P\final$ that is comparable to a networks directly trained on $P\final?$}

In other words, when the mixture contains the target distribution, can transfer learning from the mixture match the efficacy of task-specific training

\subsection{Two Ways to Train a Neural Network}

We study a network of the form 
\begin{align}
    \label{eq:linear_model}
     f(X;\theta,\bgamma) =  \gamma^T\  \varphi(X;\theta)
\end{align}
Where $\varphi(\cdot;\theta)$ is a \emph{feature extractor} (e.g. - a stack of convolutional neural networks or transformer layers) and $\gamma$ is a \emph{linear classifier}. \\
The extractor transforms the raw input $X$ into real-valued features using parameters $\theta$; the classifier produces the final prediction.

\begin{figure*}[t]
 \centering 
 \vspace{-1ex}
\includegraphics[width=0.5\linewidth]{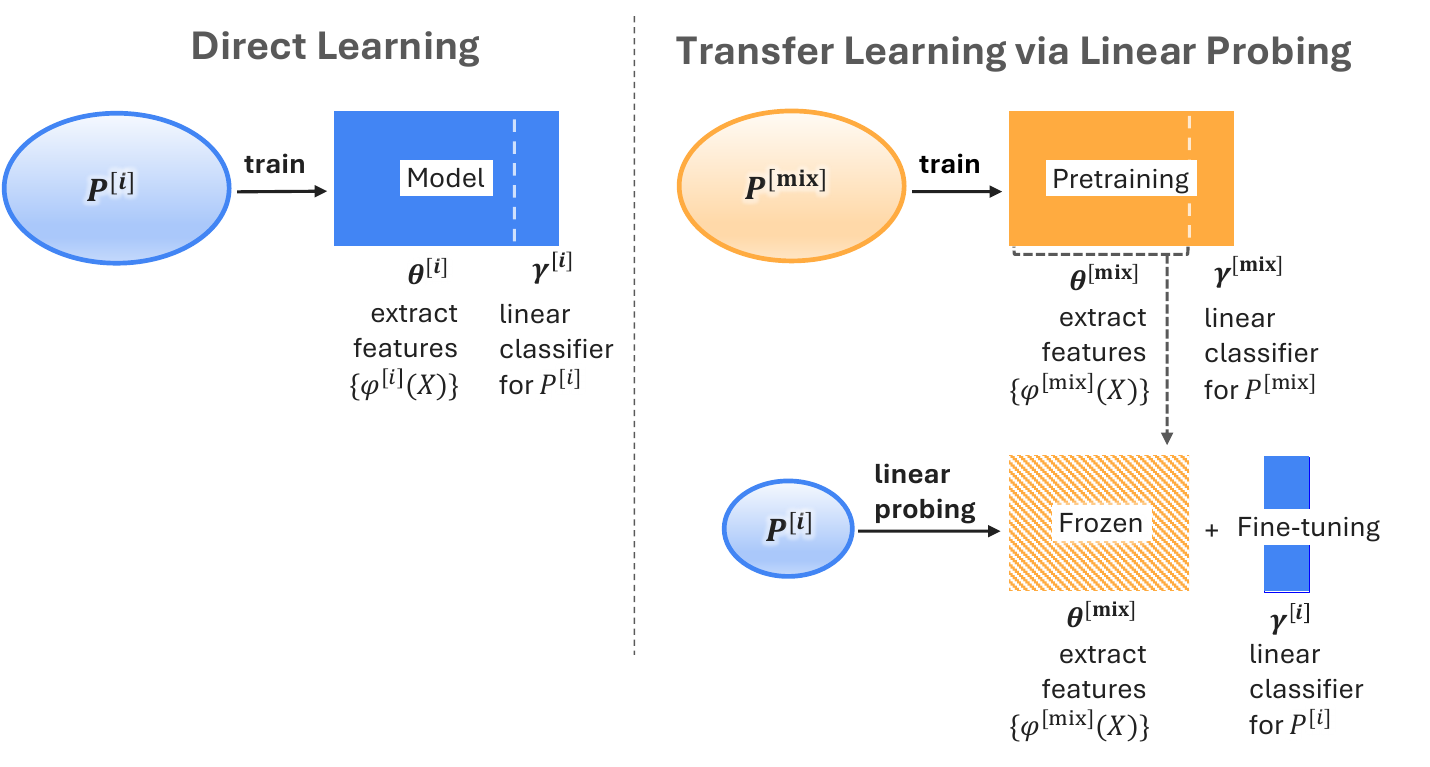}
\vspace{-1ex}
\caption{Classifier trained directly from $P\final$ versus transferred from the mixture $P\pre$.}
 \label{fig:direct-v-transfer}
\end{figure*}

There are two ways to train such a model for distribution $P\final$ (Figure \ref{fig:direct-v-transfer})
\paragraph{Direct training on a target distribution} \ \\
Given a data distribution $P\final$, we could learn both the extractor and classifier from scratch 
$$
(\theta\final,\gamma\final) = \arg\min_{\theta, \gamma} \mathbb{E}_{(X,Y\sim P\final)}[\ell(f(X; \theta,\gamma), Y)]
$$
The resulting model $f(\cdot;\theta\final,\gamma\final)$ is fully adapted to $P.$

\paragraph{Transfer learning via linear probing}
We first pretrain the extractor on a mixture distribution 
$$
P\pre(X,Y) = \sum_{j} \lambda_j P_j(X,Y), \quad \lambda_j > 0 ,\sum_j \lambda_j = 1
$$
obtaining parameters $\theta\pre$.  \\
These parameters are frozen and only the classifier is trained on $P\final$:
$$
\gamma\final_{\quad lin} = \arg\min_{\gamma}\mathbb{E}_{(X,Y)\sim P\final} [\ell(f(X;\theta\pre,\gamma), Y)].$$

The resulting "linear probe" is $f(\cdot;\theta\pre,\gamma\final _{\quad lin})$

\paragraph{The Central Question}

With unlimited data for direct training, can the performance of the directly trained model $f(\cdot;\theta\final,\gamma\final)$ be matched or approximated by the model that uses the pretrained features $\varphi(\cdot;\theta\pre)$ and only trains a linear classifier?

In other words 
\begin{align*}
&\mathbb{E}_{(X,Y)\sim P\final} [\ell(f(X;\theta\final,\gamma\final), Y)] \\
& \quad  \to (?) \mathbb{E}_{(X,Y\sim P\final)}[\ell(f(X; \theta\pre,\gamma\final_{\quad lin}), Y)]
\end{align*}

Equivalently, does the function space spanned by the pretrained mixture  $\{\gamma^T \varphi(\cdot;\theta\pre)\}$  contain the solutions  $\{f(\cdot;\theta_i, \gamma_i)\}$ obtained by training directly on the target distribution $P\final$?

Answering these questions tells us whether pretraining on a broad mixture that includes the target task suffices to capture all features needed for optimal performance on that task.

\subsection{Data Assumptions}
We take the following assumptions for a realistic model example. 
\begin{itemize} \item \textbf{Unlimited pre‑training data} – We assume an infinite stream of samples from the pre‑training distribution. Even with unlimited data, a model can still miss essential features. 
\item \textbf{Finite fine-tuning data} -- Transfer learning is applied to a small target set, which limits the adaptation of learned features. 
\item \textbf{Infinite data for linear probing} -- Fine tuned weights stay close to the pre-training weights (see \citep{radiyadixit2020finefinetuningbelearning}), so fine-tuning can be treated as linear probing on the NTK features. 
\end{itemize}

We assume unlimited training data, which ensures that a model directly trained on $P\final$ can achieve perfect generalization. 

\subsection{Linear probing is sufficient}
We first focus on linear probing, in which only the final classification layer is fine-tuned. Our analysis directly applies to this common scenario and it also covers fine-tuning because the weight updates can be expressed as a linear combination in the pre-trained features space.

\paragraph{Fine-Tuning as a linear combination} 
Let $f(X;\bw_{\pre})$ be the pre-trained network. After fine-tuning on $P\final$, the new weights $\bw\final$  remain close to the $w\pre$. A first order Taylor expansion gives 



Consider the general case of fine-tuning a pre-trained neural network $f(X;\bw\pre)$ that updates \textit{all} of its weights for target $P\final$.  
Under the assumptions of fine-tuning, the final weights $\bw\final$ remain close to the pre-trained weights $\bw\pre.$ Thus, $\bw\final$ can be reasonably approximated as a first-order Taylor expansion around $\bw\pre$
\vspace{-1ex}
\begin{align*}
   f(X;\bw\final) ~ 
   & \approx~  \underbrace{\vphantom{\frac{\partial f}{\partial _j}}f(X;\bw\pre)}_{\varphi_0(X;\ \theta\pre)}\      
\times~\underbrace{\vphantom{(\bw_j\pre)}1}_{\gamma_0}
   \quad +\\
   &\quad      \underbrace{\frac{\partial f}{\partial w_j}(X;\bw\pre)}_{\varphi_j(X;\ \theta\pre)}
   ~  \times  
    ~\sum_j 
\underbrace{(w_j\final-w\pre_j)}_{\gamma_j} 
    \space.
\end{align*}

The terms $\sum_{j}\frac{\partial f}{\partial \bw_j}(X;\mathbf{w}\pre) \bigl(w\final_{j}-w\pre_{j}\bigr)$ are exactly the  neural tangent kernel (NTK) features \citep{jacot2018neural}. Fine-tuning therefore changes only the  linear coefficients, and the problem reduces to learning a linear combination of NTK features. The same reasoning  applies if only a subset of layers (e.g. - a few non-linear classification layers) is fine-tuned.

\subsection{Models with non‑zero training error}\label{nonzero-training-error}

Training a  network until zero training error on the entire mixture would make it learn every feature, so transfer would be trivial. Such a case does not often appear in real-world settings. To study the limits of transfer in a realistic setting, we therefore consider networks that still exhibit non-zero training error after pretraining. In realistic settings -- especially with mixture distributions -- a network may overfit on high-weight components and under-fit on low-weight ones, leaving some components with non-zero test error.

\subsection{Covariance}
A feature that is correlated with the label in any component of the mixture remains correlated under almost all mixtures.



\begin{proposition}
    If $\cor_{P\final}[\varphi(X),Y] \neq 0$, for some component $P\final$, then 
    $$
    \cor_{P\pre}[\varphi(X),Y] \neq 0
    $$
    for almost all mixtures $P\pre$; only a measure-zero set of mixture proportions can cancel out the correlation. 
\end{proposition}
\begin{proof}
Assume the contrary. First, two variables are correlated if and only if they have nonzero covariance. Therefore, mixture coefficients $(\lambda_j)$ would have to satisfy
\[
    0 = 
   \sum_j \lambda_j \mathrm{Cov}^{[j]}(Y,\varphi(X)).
\]
a linear equation that is satisfied only on a measure-zero set of $\lambda_i$. Hence, the proposition holds.
\end{proof}

Thus, interestingness (non-zero correlation) transfers almost always, but correlation alone does not guarantee that a feature is learned in the solution. 



\paragraph{Sparsity in deep networks}

Stochastic gradient learning algorithms in deep learning models  often have an implicit sparsity bias \citep{gunasekar-2017,andriushchenko2023sgd}, which  likely contributes to the effectiveness of deep learning. In an oversimplification of \citeauthor{andriushchenko2023sgd}, when using SGD uses a large step sizes, it retains only features that reduce the training error; the random order of feature discovery can lead to different sparse models. 

Many practical regularization tricks such as early-stopping or weight decay have a similar effect. In particular, weight decay acts differently in the last layer of a deep network (spreading the weights over all the available features) than in the inner layers (quickly pruning features that do not rapidly help reducing the training error.)

\section{Counterexample}
We show that training on a mixture does not guarantee that learned features are as useful for each component task.

Consider a feature extractor that can produce only two real-valued functions 
$$\Phi(X)=(\varphi_1(X),\varphi_2(X))$$.
Depending on parameters $\theta,$ the extractor may use one, two, or none of these features.

\begin{figure*}[t]
    \centering
    \vspace{-1ex}
    \includegraphics[width=0.6\linewidth]{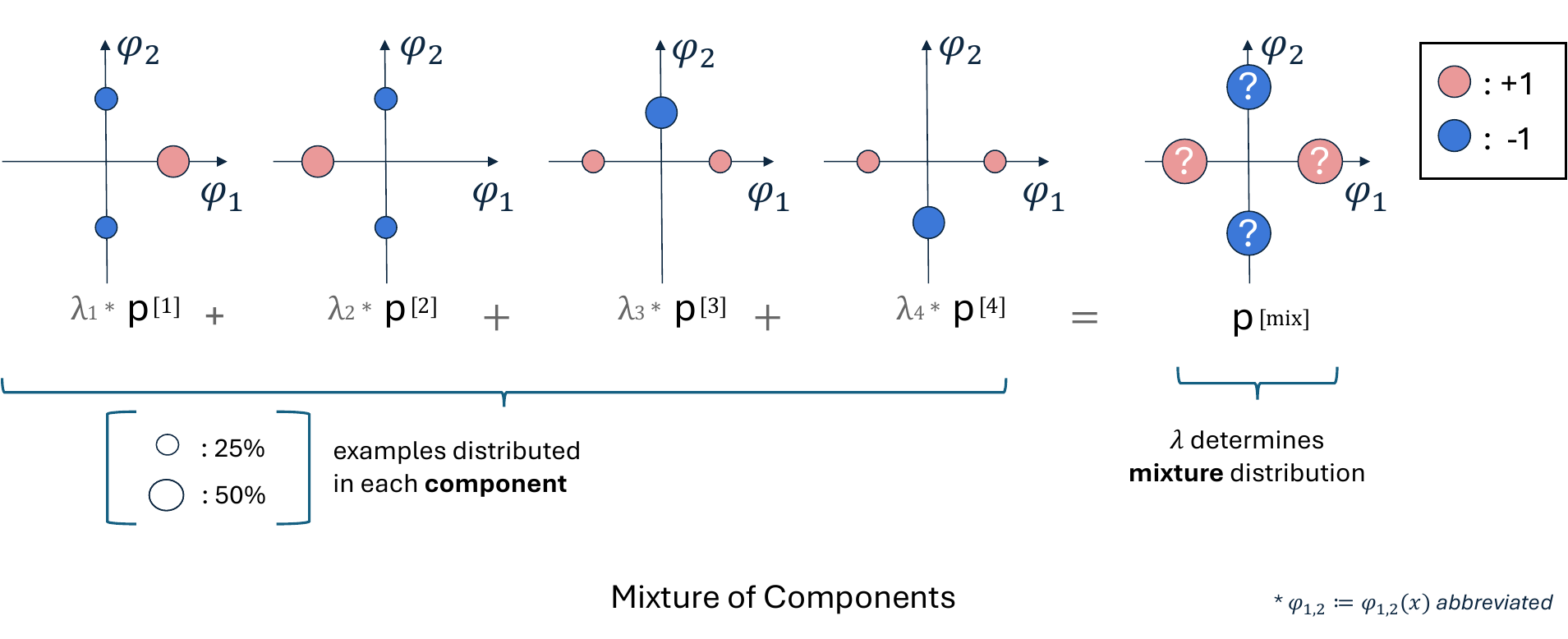}
    \caption{\textbf{Four subdistributions and a combined mixture distribution } are represented by \textcolor{red}{red} points (labeled $\color{red}+1$) and \textcolor{blue}{blue} points (labeled $\color{blue}-1$)
    Each \textit{subdistribution} includes three points; two points contain an equal number of examples, while the third point contains twice as many. The size of each point reflects the number of examples it represents.
    The final \textit{mixture distribution} is a weighted average of these four component distributions (not drawn to scale). }
    \label{fig:counter-subdist}

    \includegraphics[width=0.6\linewidth]{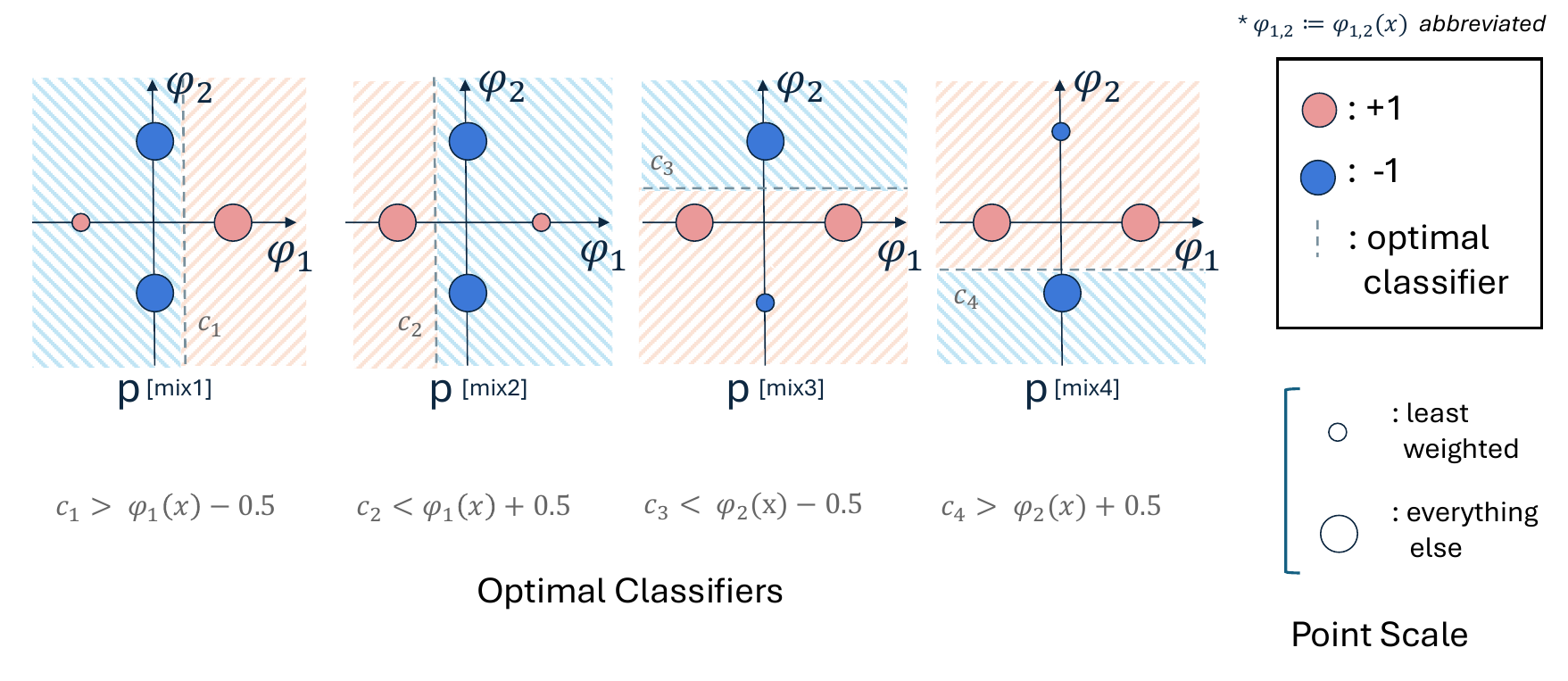}~
    \vspace{-1ex}
    \caption{Four cases illustrate \textbf{optimal classifiers} for different mixture distributions. Each  classifier uses a dotted line 
    to separate \textcolor{red}{red} points (labeled $\color{red}+1$) from the \textcolor{blue}{blue} points (labeled $\color{blue}-1$).  Since the points are not linearly separable,   an optimal classifier ignores the \textit{least-weighted point} (smallest) and classifies the remaining three points. Interestingly, this optimal classifier can be \textit{sparsely represented} using just one feature, either $\varphi_1(X)$ or $\varphi_2(X),$ but cannot classify two of the four subdistributions.
    }
    \label{fig:counter-mix}
\end{figure*}

Four toy distributions are defined on the two-dimensional space of $\Phi(x)$, illustrated in Figure \ref{fig:counter-subdist}:
\begin{itemize}
\setlength\itemsep{-0.1em}
\item $P\distrib{1}[\:\Phi(X){=}(+1,0),\:Y{=}+1\:]=\tfrac12$ and $P\distrib{1}[\:\Phi(X){=}(0,\pm1),\:Y{=}-1\:]=\tfrac14$.
\item $P\distrib{2}[\:\Phi(X){=}(-1,0),\:Y{=}+1\:]=\tfrac12$ and $P\distrib{2}[\:\Phi(X){=}(0,\pm1),\:Y{=}-1\:]=\tfrac14$.
\item $P\distrib{3}[\:\Phi(X){=}(0,+1),\:Y{=}-1\:]=\tfrac12$ and $P\distrib{3}[\:\Phi(X){=}(\pm1,0),\:Y{=}+1\:]=\tfrac14$.
\item $P\distrib{4}[\:\Phi(X){=}(0,-1),\:Y{=}-1\:]=\tfrac12$ and $P\distrib{4}[\:\Phi(X){=}(\pm1,0),\:Y{=}+1\:]=\tfrac14$.
\end{itemize}
Each distribution places a point of mass $1/2$ on a positive example and two points of mass $1/4$ on negative examples. 
The mixture
\begin{align*}
P\pre = \sum_{i= 1}^{4} \lambda_i P_i, \quad \lambda_i > 0, ~ \sum_{i=1} \lambda_i = 1
\end{align*}
has support on four points: the two positive points ${\pm 1, 0}$ and the two negative points $(0, \pm 1)$. 
See Figure~\ref{fig:counter-subdist}.

\textbf{The Optimal Classifier} \\
For each $P_i$, the optimal binary classifier uses a single feature $\varphi_1$ for $P\1$ and $P\2$ and $\varphi_2$ for $P\3$ and $P\4$. 
The mixture cannot be linearly separated; the optimal classifier misclassifies the lowest-weighted point and correctly classifies the other three. Depending on which point is ignored, the solution uses either $\varphi_1$ or $\varphi_2$. 

Since deep learning networks with sparsity bias learn only one useful feature, the mixture-trained model will choose either $\varphi_1$ or $\varphi_2$. Consequently, it can correctly classify only two of the four sub-tasks; if it learns $\varphi_1$, it fails on $P3$ and $P_4$. If it learns $\varphi_2$, it fails on $P_1$ and $P_2$, as in  Figure~\ref{fig:counter-mix}/

Thus, even in this favorable setting, pretraining on a mixture can miss essential features for its subtasks. 

\footnote{Some mixtures have multiple clusters of equal minimal weight and, therefore, multiple optimal solutions. However, this is a set of negligible probability.}

We find that training a deep network (with sparsity bias) on the mixture distribution yields a sparse representation which \emph{can classify only half of its subdistributions.}
Even in this advantageous setup, a model pretrained on a mixture may miss important features for its subtasks.

\subsection{Assumptions of the Counterexample}
In this counterexample, use the training error as a proxy for how well a model can later adapt to each subdistributions. The analysis relies on two simplifying conditions

\textbf{Assumption 1: Feature learning collapses to the problem space.}\quad \citep{papyan-2020} finds that a neural network learn representations which "collapse" into the problem space features; in our toy example, these are the two basis functions $\varphi_1(X)$, $\varphi_2(X)$ .

\textbf{Assumption 2: We Restrict Attention to Non-Linear Decision Boundaries}\quad
This type of data (XOR) may realistically appear in the NTK domain of a learned model. When fine-tuning the last layer features of this model on a smaller target set, this mirrors linear probing of the NTK features (as explained L108-117). There is no way to fine tune or introduce non-linearities within the NTK feature space to overcome this limitation.




\section{Evidence from Existing Papers}

Our counterexample shows that a model can fail to learn essential features, leading to biased performance on sub‑tasks of the training distribution. Even when optimally trained, a network may omit a critical feature in this toy setting—suggesting that deep networks often overlook important features when applied to more complex data.

\subsection{Hypothesis: Sparsity Bias Limits Feature Learning}

We argue that deep neural networks possess an inherent sparsity bias--arising from SGD, weight decay, and early stopping--that restricts the set of features that they can learn.

When a network has encoded a feature that is similar to a new one, the optimizer tends to keep the existing feature and discard the new one. Consequently, only a subset of the truly informative features is retained, and critical features that would help later transfer are lost. 

In our counterexample this manifests as inconsistent performance on different components of the pretraining distribution. The order in which features are learned depends on arbitrary factors such as data shuffling and class‑balance, which introduces a bias in the resulting representation.

This perspective explains several previously reported phenomena; rather than repeating experiments, we describe the limitation and survey relevant work below.

\subsection{Example - Spurious Features} 


\begin{table*}
\centering
\vspace{-2ex}
\caption{\textbf{Empirical Evidence} \emph{(Left)} Performance gap after recovering core features with transfer
\emph{(Right)} The simple supervised models trained by their supervised 'DASHA' workflow outperform eight well Genomic Foundation Models (GFMs) on the diverse NT benchmark \citep{xu2025specializedfoundationmodelsstruggle}}
\vspace{-1ex}
\setlength{\tabcolsep}{2pt}
\begin{tabularx}{\linewidth}{@{}c X @{}}

\resizebox{0.4\textwidth}{!}{
 \begin{tabular}{c c c c r } 
 \toprule
\multicolumn{1}{c}{Trial} && \multicolumn{3}{c}{Accuracy} \\ 
\cmidrule{1-1} \cmidrule{3-5} 
\textit{Data} &&\textit{Direct} &  \textit{Transfer via LinProb } &  \textit{Change}\\
($99\%$ corr)&& $P\bal$ &   $P\pre\to P\bal$ &  \\
\midrule
MNIST-Fashion&&  $97\%$&  $94\%$ &($\color{red}-3\%$)   \\  
MNIST-CIFAR &&\  $90\%$ & $81\%$ & ($\color{red}-9\%$)\\ 
\bottomrule \\

\end{tabular} 
\label{table:iid-linprob-kirichenko}

}& 
        \resizebox{0.6\textwidth}{!}{
        \begin{tabular}{cl|cc|cccc}
        \toprule
        &\multirow{2}{*}{Model}   & Model &Pretraining & Average & Average &Mean&Median\\ 
        &&Size &Base-Pairs& Score $\uparrow$ & Rank $\downarrow$ & \%Imp.$\uparrow$&\%Imp.$\uparrow$\\
        \midrule
        &Enformer &  252M & 4B & 0.569 & 11.86 & 27.73 & 27.91 \\
        &NT-1000G (500M)  & 500M&20.5T& 0.625 & 10.52 & 33.48 & 36.74 \\
        &NT-1000G (2.5B) & 2.5B&20.5T& 0.656 & 7.0 & 36.58 & 40.86 \\
        &NT-Multispecies (500M) & 500M &174B& 0.700 & 3.81  & 40.76 & 45.07 \\
        {\bf Foundation}
        &NT-Multispecies (2.5B) & 2.5B &174B& 0.697 & 4.08  & 40.51 & 45.52 \\
        {\bf Models}
        &DNABERT-2 & 117M & 32.5B & 0.680 & 6.88  & 38.65 & 43.59 \\
        &HyenaDNA-1K & 1.6M &3.2B& 0.708 & 6.92  & 41.2 & 43.36 \\
        &HyenaDNA-32K & 1.6M &3.2B& 0.630 & 10.22  & 33.96 & 36.93 \\
        &Caduceus-PS & 1.9M & 35B & 0.689 & 6.69 & 39.08 & 41.38\\
        &Caduceus-PH & 1.9M & 35B & 0.725 & 4.69 & 42.63 & 45.01\\
        \midrule
        \multirow{2}{*}{\bf Supervised}
        &Wide ResNet &2.0M & 0 & 0.694 & 6.83  & 37.16 & 43.08 \\
        \multirow{2}{*}{\bf Models}
        &UNet &4.5M & 0 & 0.68 & 7.78  & 38.67 & 42.69 \\
        &\textbf{DASHA (our workflow)} &10.5M & 0 & \textbf{0.761} & \textbf{3.69}  & \textbf{46.33} & \textbf{49.08} \\
        \bottomrule
        \end{tabular}}
        \label{tab:empirical-gfm-results}
        

\end{tabularx}

\vspace{-4ex}
\end{table*}
Spurious features are superficial features which are \emph{correlated with} but \emph{do not predict} the label. For example, a bird classification model may incorrectly learn to use a water background to predict a waterbird and a land background to predict a land bird.

\citep{pezeshki2021gradient} find that learning spurious features hinders the appearance of \textit{core features} essential for accurate prediction. They find that spurious features to arise from bias towards certain classes in complicated mixtures of data, descriptive of real-world datasets. Core features cannot appear after the gradient is consumed by spurious features, which provide the same information but cannot be used for prediction. Class imbalances produce spurious features which "starve" the model of essential features.

\citep{kirichenko2023last} propose a promising mitigation to recover core features via transfer, by reweighting the remaining features without the spurious feature $\varphi_{\text{spu}}$. They consider that a model, pretrained on a biased mixture $P\pre$, may still learn \emph{core features} after learning easier spurious features. They remove $P\varphi_{\text{spu}}$ by transferring the network to a class-balanced subdistribution $P\bal$ (where $\cor\big(\varphi_{\text{spu}}(X), Y\big) = 0$). The remaining core features are optimized via linear probing.

Despite significant accuracy improvements, a performance gap $\{3\%, 9\%\}$ remains between the adapted model and one trained directly on $P\bal$ (Table \ref{table:iid-linprob-kirichenko}). Here, directly training on P$\bal$ produces a solution which includes \textit{all} of the core features. In contrast, pre-training on $P\pre$ gives a solution which does not. Linear probing on $P\bal$ cannot recover these features.

In this example of the bottleneck, \textit{pretraining on an imbalanced mixture} produces \textit{misleading spurious features}, which prevent the network from learning core features which are essential for prediction.

\subsection{Example - Genomic Foundation Models}
\citep{xu2025specializedfoundationmodelsstruggle} find that simple supervised CNNs can be easily trained to match the performance of transformer-based genomic foundation models (GFMs). Using their "DASHA" workflow, they train and evaluate supervised models on a diverse set of genomic tasks $P\adja$ from the Nucleotide Transformer (NT) benchmark.
They compare to the performance of well-known GFMs with up to 2.5B parameters, which are pretrained on a broad mixture $P\pre$ of genomic tasks.
They find the directly trained models to \textit{consistently outperform} the GFMs across all aggregated metrics of $P\adja$ benchmark (Table \ref{tab:empirical-gfm-results}).

Certain GFMs such as Caduceus or NT-Multispecies(500M) perform almost as well as supervised networks on specific tasks of $P\adja$, but miss features needed to perform well across all $P\adja$. This suggests that large scale models learn some, but not all critical features for $P\adja$ which appear in the extensive $P\pre.$  While GFMs can excel in specific scenarios, their performance on new tasks is, ultimately, constrained by missing features that arise from information saturation.

If foundational models saturate regardless of scale, it may be more efficient invest resources into creating direct training data.

\subsection{Example - Rich Representations: The Beginnings of a Solution?}
\citep{zhang2023learning} propose learning \textit{richer representations} to capture these missing features. They  concatenating feature representations of individual models and demonstrate significant improvements in accuracy of transfer.

To construct rich representations, they concatenate four copies of \textsc{ResNet50} trained separately on \textsc{ImageNet1K}, with different random seeds, into an \textit{ensemble} model (\textsc{ResNetCat4}). 
They evaluate against a \emph{baseline} model (\textsc{ResNet50W2}) of the same increased size, trained once on \textsc{ImageNet1K}.

\begin{figure*}
\centering \includegraphics[width=0.7\linewidth]{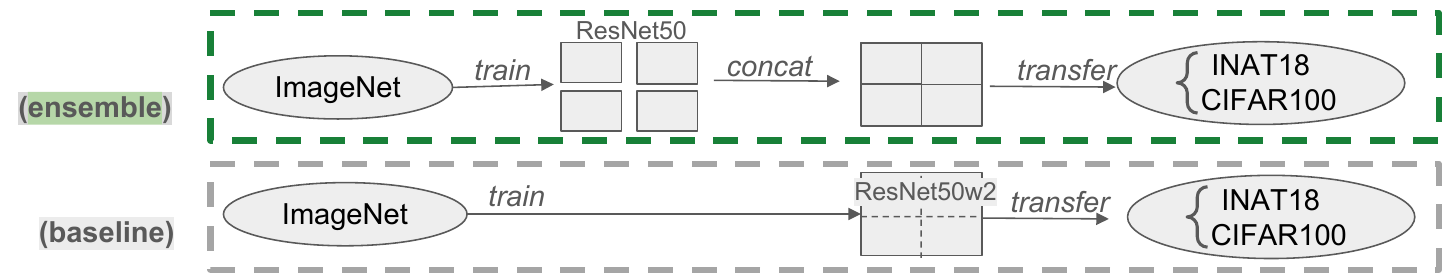}
\caption{\textbf{Richer Representations via Concatenation} An \emph{ensemble} model (ResNet50Cat4) is created by concatenating four ResNet models trained separately on ImageNet using different random seed. It is evaluated against a \emph{baseline} model (ResNet50W2), which is a single ResNet model of the same size, trained once on ImageNet.}
\label{fig:rich-reps-cat-diagram} 
    \vspace{-1ex}
    \vspace{-1ex}

\end{figure*}
\begin{table}[b]

    \centering
    \caption{\textbf{Supervised transfer learning from \textsc{ImageNet} to \textsc{iNat18}, \textsc{Cifar100}, and \textsc{Cifar10} using linear probing.}
    The \textit{\textsc{cat4} ensemble} concatenates representations of $4$ separately trained networks. The \textit{ baseline} uses a ResNet model of similar size is trained once. These models contain approximately four times the parameters of the \textit{original}  ResNet50.}

    \resizebox{0.50\textwidth}{!}{

    \begin{tabular}{c cc  |c| ccc }
    \toprule

                 &              &        &    {\small Direct}             &      \multicolumn{3}{c}{ Transfer via LinProb} \\
         Method  & Arch     & Params & \textsc{\small ImageNet1k} &  \textsc{\small iNat18}  & \textsc{\small Cifar100} & \textsc{\small Cifar10} \\
         \midrule
             {original} & \textsc{resnet50}     & 23.5\textsc{m} & 75.58 & 37.91 & 73.23 & 90.57   \\
             \colorbox{lightgray}{baseline} & \textsc{resnet50w2}   & 93.9\textsc{m} & 77.58	& 37.34 & 72.65 & 90.86	\\
             \colorbox{LimeGreen}{ensemble} & \textsc{cat4}$\times$\textsc{resnet50}     & 94\textsc{m}   & 78.15 & 46.55	& 78.19	& 93.09	\\
        \bottomrule

    \end{tabular}
    }
    \label{tab:imagenet_sl_lineareval}
    
\end{table}

They find the concatenated ensemble to perform comparably to the baseline on \textsc{ImageNet1K}, but \emph{far better} when transferred to new datasets, ($+9\%$ on \textsc{INat18}) and ($+5.5\%$ on \textsc{CIFAR100}). Additional comparison to the original \text{ResNet50} reveals that only increasing size of a model (such as the baseline) will worsen performance of transfer: ($-.5\%$ on \textsc{INat18}), and  ($-.6\%$ on \textsc{CIFAR100}). 

The performance gain of the concatenated ensemble can only be attributed to the features recovered by a richer, combined representation. This means that the four concatenated networks \textit{learn different subsets of features when trained on identical data shuffled differently.}

This also explains why simply increasing model size without also increasing model richness produces similar or slightly worse results.  Observe that these subsets of features work similarly on ImageNet, that is, each set of features provides the same information to reduce training error. Combining these features \textit{marginally} improves performance on \textsc{ImageNet1K} ($+2\%$). Hence, they are not learned during training.

Using a similar concatenation approach, \citep{zhang2022rich} demonstrate increased returns in accuracy for Self-Supervised Learning as scale increases from 100M to 1B parameters, as well as on Vision Transformers as scale increases from 100M to 400M parameters (Figure \ref{fig:rich-reps-cat-diagram}). In other words, these rich representations recover crucial features that otherwise would be lost and significantly increases performance accuracy without incurring additional training costs.

\begin{figure*}
\centering
\includegraphics[height=0.22\linewidth]{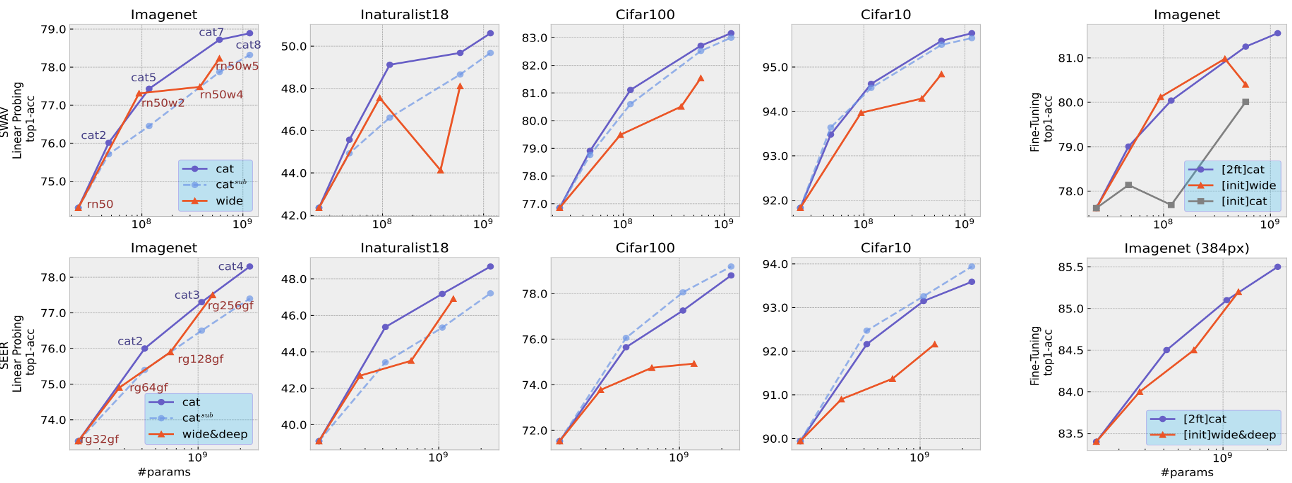}\quad\quad
\includegraphics[height=0.22\linewidth]{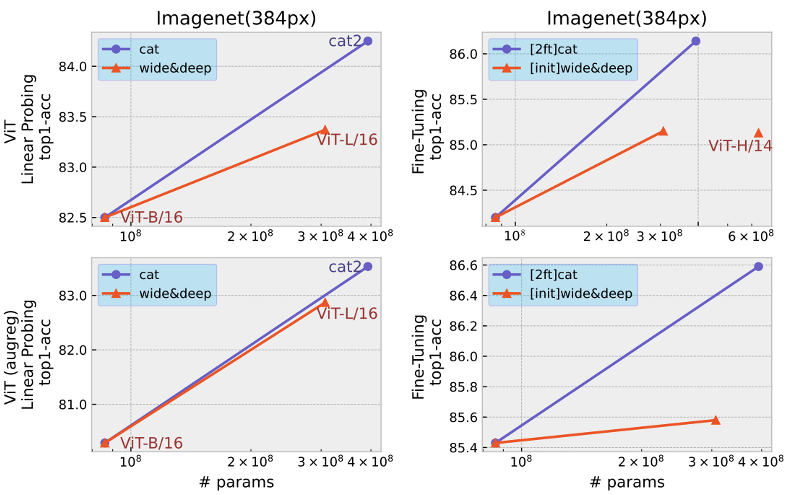}
\caption{\textbf{Increased Returns in Accuracy as Scale Increases} \textit{(Left)} \textbf{SSL, 100M to 1B params:} 
The concatenated methods (in \textcolor{red}{red} and \textcolor{violet}{purple}) outperform the baseline (\textcolor{cyan}{dotted blue} curve) on both \textsc{SWAV} trained on unlabeled \textsc{ImageNet1K} (top), and \textsc{SEER} on \textsc{INSTAGRAM1B} (bottom). \textit{(Right)} \textbf{ViT, 100M to 400M params:} 
Concatenated representations (\textcolor{violet}{purple}) outperform the baseline (\textcolor{red}{red}) consistently during transfer in both original (top) and modified (bottom) vision transformers (\textsc{ViT})}
\label{fig:rich-reps-ssl}
\end{figure*}

\section{Our New Experiments}
\label{sec:experiments}

\subsection{Improved Transfer at Fixed Cost}

We propose a method to improve transfer performance with fixed total compute. Our method creates rich representations by ensembling, following the method of \citep{zhang2022rich} to concatenate networks pretrained for fewer steps.

In contrast to \citep{zhang2022rich}'s experiments which preserve model parameters (space), our method fixes total compute (\emph{time}). We call our method (\textsc{time-concat}).

\subsubsection{Methodology}
Total pretraining time is fixed at 450k iterations per ResNet50 model on 8xH100 GPUs on ImageNet1k \citep{imagenet_cvpr09}, \citep{he2016deep}.%
\footnote{Implementation can be found here: \url{https://github.com/facebookresearch/richreps-timecat}}
All pretraining uses the AdamWScheduleFree Optimizer and builds on the MLCommons AlgoPerf Training Framework \citep{Kasimbeg2025AlgoPerfResults}, \citep{defazio2024roadscheduled}.

\begin{itemize}
    \item Baseline Model: A single ResNet50 model fully-trained on ImageNet1k.
    \item Ensemble Model: Multiple ResNet50 models trained independently for shorter periods (different seeds, then ensembled).
\end{itemize}
For example, an ensemble could contain four models trained for $25\%$ of the total time or two models trained for $50\%$ of the total time.

Each \textsc{ResNet50} network is trained identically except using different random seeds. We transfer these concatenated networks using linear probing on new datasets.

\subsubsection{Evaluation}

Models are transferred to datasets via linear probing (fine-tuning only the final classification layer) and evaluated on new datasets.

In-Distribution (IID) Performance is measured after transfer to the pretraining dataset (ImageNet1k).

Out-of-Distribution (OOD) Performance is evaluated on datasets not seen during pretraining (\textsc{iNaturalist18}, \textsc{CIFAR-100}, \textsc{CIFAR-10}) \citep{van2018inaturalist}, \citep{krizhevsky2009learning}.

\begin{figure*}[t]
    \centering
    \includegraphics[width=0.75\linewidth]{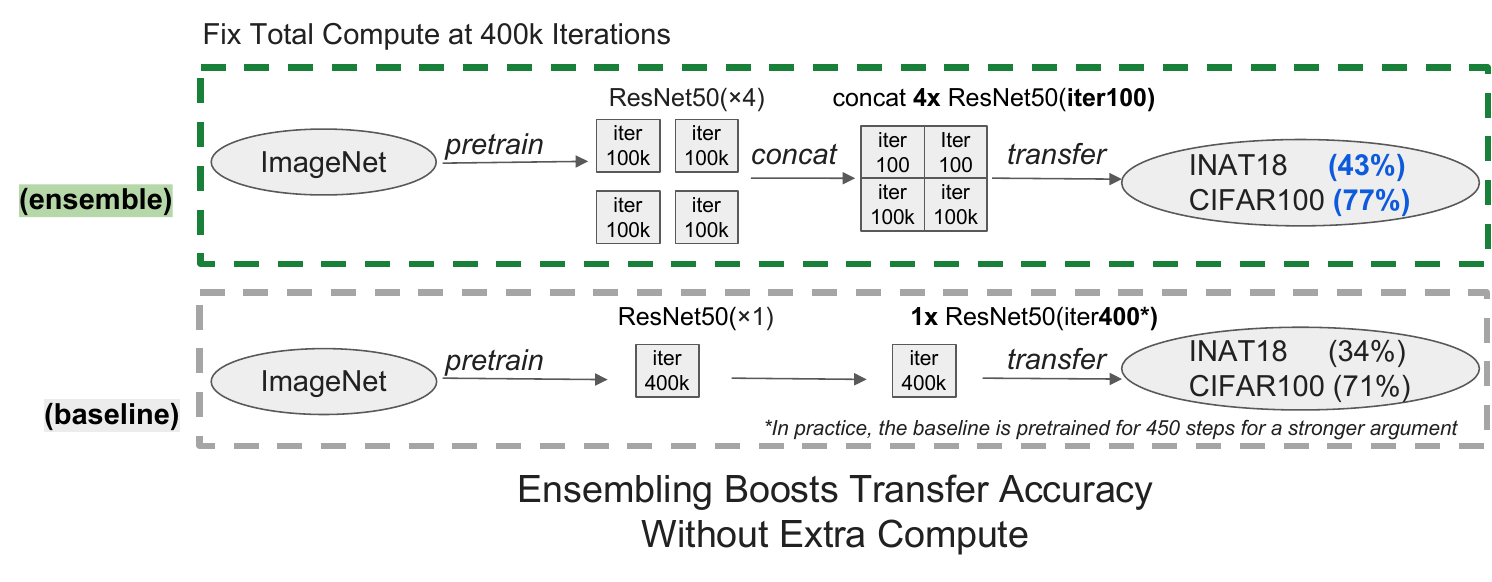}
    \caption{\textbf{Ensembling Boosts ResNet50 Transfer Accuracy Without Extra Compute}
    \emph{Baseline:} A single (\textsc{ResNet50Cat1}) model is trained for 400k iters on ImageNet1k (plus $50$k iters for a stronger argument).
    \emph{Ensemble:} A (\textsc{ResNet50Cat4}) model contains four \textsc{ResNet50} models trained separately (with different random seeds) on ImageNet for 100k iters each. The ensemble significantly outperforms the baseline during transfer.}
    \label{fig:resnet-concat-diagram}
\end{figure*}
\footnotetext{The baseline is pretrained for 450k training steps for a stronger argument.}

\subsubsection{Results}

Our method yields considerable improvements in performance at fixed training cost.

On ImageNet1k (IID), the ensemble models performed comparably or slightly worse than the single baseline model, showing no difference in their knowledge of ImageNet1k. However, on OOD (Green), \emph{the undertrained ensemble significantly outperforms the fully-trained baseline} ($+12.5\%$ on \textsc{iNaturalist18}, $+6\%$ on \textsc{CIFAR-100}).

This performance gap suggests that an ensemble of undertrained models captures patterns relevant to new tasks which a single fully-trained model may miss.

In summary, for a fixed computational budget, an ensemble of intentionally under-trained models outperforms a single, fully-trained model when transferred to unseen distributions.

As expected, concatenating additional sets of features does not significantly boost model performance on \textsc{ImageNet1K}. \textsc{ResNet50} models generate different sets of features which provide similar information to reduce training loss. Thus, combining them on \textsc{ImageNet} yields no further improvements.

If concatenating more models trained on fewer steps can yield a significantly better performing model on new datasets, there may be considerable value left on the table.

See Table~\ref{tab:concat_fixed_time_resnet} for detailed results.

\begin{table}[h]
    \caption{\textbf{Transfer Performance of Concatenated ResNet models with Fixed Total Compute}
    A \emph{ baseline} \textsc{ResNet50} model (\textsc{cat1}) is trained for \textsc{450k} iters on ImageNet1k.
    Each experimental method (\textsc{catn}) creates a models by concatenating multiple (\textsc{n}) \textsc{ResNet50} networks, each trained independently for \textsc{400k}/\textsc{n} iters. Overall computational effort of pretraining is maintained.}
    \par\vspace*{-1ex}
    \label{tab:concat_fixed_time_resnet}
    \centering
    \resizebox{.5\textwidth}{!}{
    \begin{tabular}{r cc  | cccc }
    \toprule

          \multicolumn{3}{c}{ Pretraining Iters} &    \multicolumn{3}{c}{Transfer via LinProb} \\
         Method  &   Per ResNet50    & Total  &   \textsc{\small ImageNet1k} &   \textsc{\small Inat18}  & \textsc{\small Cifar100} & \textsc{\small Cifar10} \\
         \midrule
             (BL) \textsc{cat1}& 450\textsc{k}     
                & 450\textsc{k} & {78.4\%} & {33.7\%} & 71.0\% & 90.4\%   \\
             \textsc{cat2}& 200\textsc{k}\     
                & 400\textsc{k} &  {\bf78.5\%} & {40.4\%} & 74.2\% & 91.2\%  \\
             \textsc{cat4}& 100\textsc{k}\     
                & 400\textsc{k}  &{78.3\%} & {42.9\%} & $76.8\%$ & 92.7\%   \\
             \textsc{cat5}& 80\textsc{k}\     
                & 400\textsc{k} & {77.8\%} &{\bf46.2\%} & \bf77.1\% & \bf92.8\%   \\
        \bottomrule
    \end{tabular}
v    } 
\end{table}

\section{Limitations}
This paper identifies a behavior in deep neural networks, but does not claim to fully discover, solve, or explain this. 
First, this paper acknowledges prior work on related issues, such as spurious features from imbalanced datasets or poor generalization from foundation models.
This paper builds on previous research by using their experimental results as a foundation for its own findings.  While it presents rich representations as a potential solution, this paper also recognizes that this solution is incomplete. Instead, it serves as the initial steps towards understanding the problem.
Finally, this paper focuses on specific toy examples and empirical evidence to provide insight into when and how this behavior might occur. 
These results are not comprehensive but provide an early phase of understanding, beginning with the classical case of fine-tuning. 

\section{Conclusion}
This paper explores whether training models across a mixture of tasks provides the necessary features for each component. Our results indicate that supervised transfer methods, however useful, often permanently lose essential features without justification, potentially limiting their ability to generalize effectively. External factors like dataset imbalance or random seed limit what subpopulations a model can learn before training starts. Foundation models cannot alleviate this phenomenon with scale alone. We identify the beginnings of a solution to retrieve these lost features, revealing a fresh problem space with significant potential.
\newpage
\bibliographystyle{plainnat}
\bibliography{neurips_mixtures}
\newpage
\section{Appendix}
\appendix

\section{Additional Empirical Evidence}
This paper introduces a phenomenon called the $\textit{information saturation bottleneck}$ where models saturate information from encoded features and, consequently, miss key features. This can cause models to perform inconsistently across different subgroups of data.

This bottleneck appears not just in the cases discussed in this paper, but is widespread  across the existing machine learning literature. Below are a few additional examples where this issue has been observed.

\paragraph{Biological Foundation Models}
Beyond the work in this paper, a number of other works use supervised model to match the performance of larger-scale models.
\begin{itemize}
\item Convolutional method can match the performance of Transformer-based foundation models \citep{Yang2022.05.19.492714}
\item Domain-specific generative model can outperform pretraining for single-cell biology tasks \citep{Kedzierska2023.10.16.561085}
\end{itemize}

\paragraph{Fairness}
\begin{itemize}
\item Fairness treatment of individuals within classification tasks \citep{dwork2012fairness}
\item Identifying biased treatment (discrimination on sensitive attributes) in supervised learning \citep{hardt2016equality}
\item Incompatibility between fairness conditions and accuracy of risk scores
\citep{kleinberg2016inherent}
\end{itemize}
\paragraph{Robustness}
\begin{itemize}
\item  Data augmentation for the ImageNet task  \citep{data_augmentation}, \citep{kirichenko2023understandingdetrimentalclassleveleffects}
\item Group-DRO on water-bird, CelebA, MultiNLI tasks 
\citep{sagawa-2019}
\end{itemize}

\section{Connections to Multi-Calibration}

Missing features fundamentally limits the ability of a model to generalize to new distributions outside the training distribution (OOD). \citep{wald2021calibration} finds a link between OOD performance and multi-calibration, introduced by \citep{hebertjohnson18a}. Specifically, they argue that calibration leads to better OOD generalization, by creating invariant representations \citep{arjovsky2019invariant}. \citep{guy2025measuringmulticalibration} successfully demonstrate this empirically for in-distribution generalization (IID). However, any remediation requires recovering these missing features.

\end{document}